\documentclass[conference]{IEEEtran}
\IEEEoverridecommandlockouts
\usepackage{graphicx}
\usepackage{booktabs}
\usepackage{amsmath}
\usepackage{microtype}
\usepackage{url}
\usepackage{balance}
\usepackage{amssymb}
\usepackage{xcolor}
\newcommand{\cmark}{\textcolor{green!60!black}{\checkmark}}
\newcommand{\xmark}{\textcolor{red!75!black}{\ensuremath{\times}}}
\usepackage{colortbl}

\definecolor{oursgreen}{RGB}{235,247,235}

\title{RAVEL-KV: Request-Level Analysis of Violation Events in KV-Cache Compression}

\author{\IEEEauthorblockN{Saim Rehman, Muhammad Shafique}
\IEEEauthorblockA{eBRAIN Lab, Division of Engineering, New York University Abu Dhabi\\sr7849@nyu.edu, muhammad.shafique@nyu.edu
}}

\begin{document}
\maketitle

\begin{abstract}
Key--value (KV) cache compression reduces the memory cost of long-context large language model (LLM) inference, yet average task scores can obscure request-level failures. We present RAVEL-KV, a controlled request-level characterization of heterogeneous KV-cache compression. We evaluate 27,951 Mistral-7B-Instruct v0.2 generations over 1,331 LongBench requests, nine tasks, four compression mechanisms, and five retained-token budgets. At a FullKV-relative quality-loss tolerance of 0.10, 16.26\% of compressed request--configuration pairs violate the threshold, with failures strongly dependent on mechanism and budget. A lightweight predictor identifies these violations before compressed execution, achieving $0.8719\pm0.0026$ AUROC and $0.5941\pm0.0098$ AUPRC across five repeated request-grouped evaluations. A request-wise oracle achieves 94.17\% mean nominal retained-token saving with zero violations, revealing substantial adaptive-execution headroom. These results motivate treating KV-cache compression as a request-level risk problem rather than solely an average-quality problem.
\end{abstract}

\begin{IEEEkeywords}
Large Language Models, KV Cache, Long-Context Inference, Compression,
Runtime Efficiency, Quality Risk
\end{IEEEkeywords}

\section{Introduction}
Autoregressive transformers reuse the key and value tensors of preceding tokens
to avoid recomputing attention states during generation \cite{vaswani2017attention}.
For long contexts, however, the KV cache can become a dominant memory and
memory-bandwidth cost. Systems such as PagedAttention address KV-memory
management \cite{kwon2023pagedattention}, while a rapidly growing line of work
compresses or sparsifies the cache itself through token eviction, head-wise
allocation, cross-layer merging, selective retrieval, or quantization
\cite{zhang2023h2o,liu2023scissorhands,li2024snapkv,cai2025pyramidkv,
feng2025adakv,liu2024kivi,liu2024minicache,tang2024quest}.

Most such methods are reported as a small collection of average quality points
at one or more compression ratios. This view is useful for comparing methods,
but it can conceal a deployment-critical phenomenon: the same compression
setting can be harmless for one request and destructive for another. This
matters because long-context applications mix question answering, retrieval,
summarization, classification, and aggregation workloads whose dependence on
past tokens differs substantially \cite{bai2024longbench,hsieh2024ruler}.
Recent adaptive methods already demonstrate that task-, layer-, head-, or
query-dependent information can improve cache allocation
\cite{feng2025adakv,zhou2025dynamickv,tang2024quest,xiao2025duoattention}.
More recently, MetaKV selects compression configurations per prompt under
latency and memory constraints \cite{wang2026metakv}. These developments make
the request-level behavior of heterogeneous cache configurations increasingly
important.

RAVEL-KV asks a narrower question: \emph{how heterogeneous is
compression-induced quality loss at the request level, and is the risk of a
large quality loss predictable before executing a compressed candidate?}
\textbf{Our three novel contributions are as follows:}
\begin{itemize}
\item A matched evaluation of 27,951 generations covering 1,331 requests, nine
LongBench tasks, FullKV, and 20 compressed method--budget configurations.
\item A request-level degradation analysis showing that equal retained-token
budgets are not equal quality operating points across mechanisms.
\item Evidence that quality-loss violations are predictable under
request-grouped validation, reaching $0.8719\pm0.0026$ AUROC at a 0.10
degradation threshold; a request-wise oracle reveals 94.17\% mean nominal
retained-token saving with zero violations.
\end{itemize}

The scope is intentionally characterization and predictability. We do not claim
measured GPU-memory savings from nominal token retention, and we leave
risk-constrained runtime synthesis, resource telemetry, and closed-loop
adaptation to separate systems work.

\begin{figure*}[t]
    \centering
    \includegraphics[width=1\linewidth]{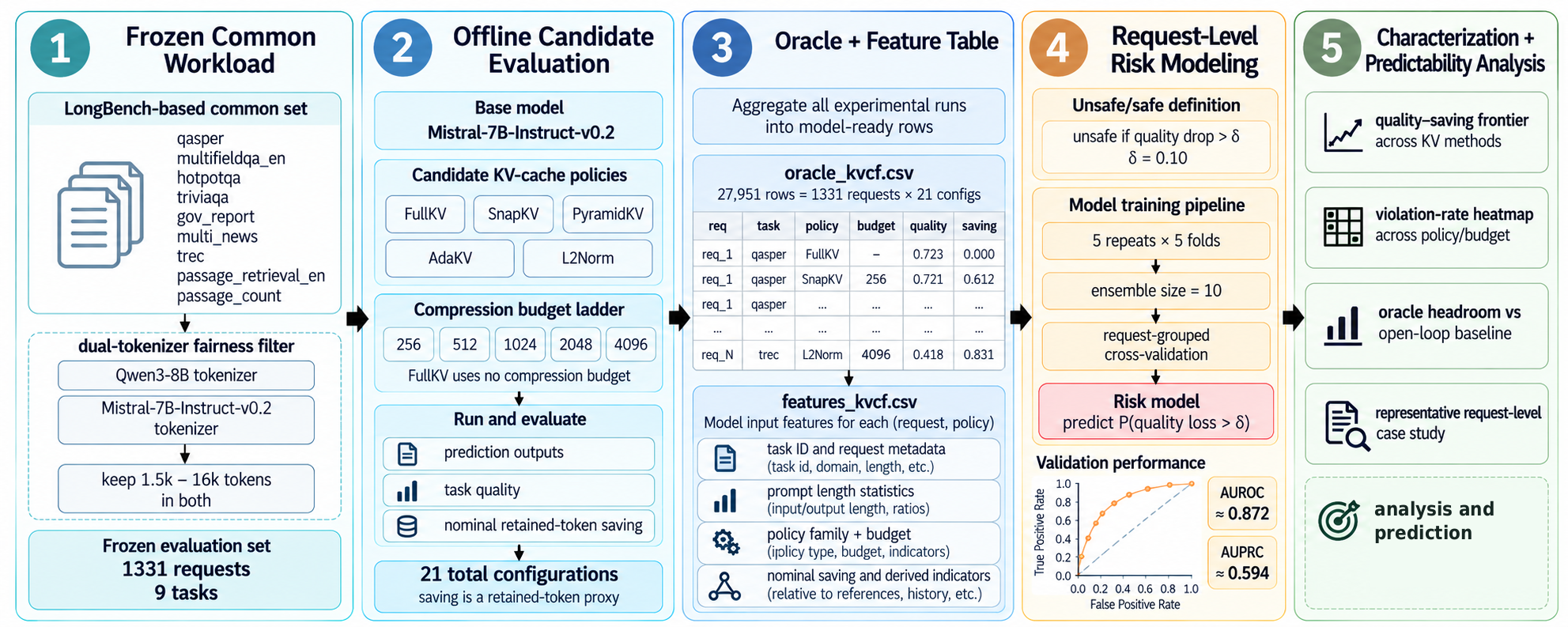}
    \caption{Analysis pipeline for the study: frozen workload construction, matched offline KV-cache evaluation, FullKV-relative request labeling, grouped risk modeling, and characterization/predictability analysis.}
    \label{fig:methodology}
\end{figure*}

\section{Related Work}
\textbf{KV-cache eviction and sparsity.}
H$_2$O identifies heavy-hitter tokens and preserves them under a cache budget
\cite{zhang2023h2o}. Scissorhands similarly exploits persistence in token
importance \cite{liu2023scissorhands}. StreamingLLM shows the importance of
attention sinks when using windowed caches \cite{xiao2024streaming}. SnapKV
uses an observation window to identify prompt positions that remain important
during generation \cite{li2024snapkv}, while PyramidKV varies cache allocation
across layers according to pyramidal information flow \cite{cai2025pyramidkv}.
Ada-KV allocates eviction budgets across heads \cite{feng2025adakv}. Quest and
SparQ reduce attention data movement through query-aware or selective cache
access \cite{tang2024quest,ribar2024sparq}. DuoAttention distinguishes
retrieval heads from streaming heads \cite{xiao2025duoattention}, and RocketKV
uses two-stage cache compression \cite{behnam2025rocketkv}.

\textbf{Quantization, cross-layer compression, and adaptation.}
KIVI develops asymmetric 2-bit KV quantization \cite{liu2024kivi}; MiniCache
compresses across adjacent layers \cite{liu2024minicache}. DynamicKV adapts
layer-wise retained tokens to task-dependent activation patterns
\cite{zhou2025dynamickv}. MetaKV predicts candidate correctness, latency, and
peak memory to perform per-prompt constrained selection \cite{wang2026metakv}.
Our study is complementary: rather than proposing another compression
mechanism or a complete selector, we isolate request-level quality risk across a
matched heterogeneous candidate set.

\begin{table}[t]
\centering
\caption{Comparison of representative KV-cache approaches and the analysis dimensions emphasized in RAVEL-KV.}
\label{tab:method_comparison}
\renewcommand{\arraystretch}{1.10}
\setlength{\tabcolsep}{1.8pt}
\scriptsize
\resizebox{\columnwidth}{!}{%
\begin{tabular}{lccccc}
\toprule
\textbf{Method}
& \shortstack{\textbf{Adaptive}\\\textbf{Compression}}
& \shortstack{\textbf{Heterogeneous}\\\textbf{Mechanisms}}
& \shortstack{\textbf{FullKV-Rel.}\\\textbf{Loss}}
& \shortstack{\textbf{Request-Level}\\\textbf{Violations}}
& \shortstack{\textbf{Pre-Exec.}\\\textbf{Violation Pred.}} \\
\midrule
H$_2$O \cite{zhang2023h2o}          & \cmark & \xmark & \xmark & \xmark & \xmark \\
SnapKV \cite{li2024snapkv}          & \cmark & \xmark & \xmark & \xmark & \xmark \\
PyramidKV \cite{cai2025pyramidkv}   & \cmark & \xmark & \xmark & \xmark & \xmark \\
Ada-KV \cite{feng2025adakv}         & \cmark & \xmark & \xmark & \xmark & \xmark \\
DynamicKV \cite{zhou2025dynamickv}  & \cmark & \xmark & \xmark & \xmark & \xmark \\
MetaKV \cite{wang2026metakv}        & \cmark & \cmark & \xmark & \xmark & \xmark \\
\midrule
\textbf{RAVEL-KV}                  & \xmark & \cmark & \cmark & \cmark & \cmark \\
\bottomrule
\end{tabular}%
}
\vspace{0.5mm}
\raggedright\footnotesize
\cmark: explicitly studied/supported; \xmark: not a primary component.
\end{table}

\section{Methodology}

\subsection{Model, Tasks, and Candidate Set}
Experiments use Mistral-7B-Instruct-v0.2 \cite{jiang2023mistral}. We select
1,331 English LongBench requests \cite{bai2024longbench} from nine tasks:
Qasper (196), MultiFieldQA-en (143), HotpotQA (89), TriviaQA (123),
GovReport (161), MultiNews (147), TREC (200), PassageRetrieval-en (186), and
PassageCount (86). To avoid comparing different truncation regimes, requests
are retained only when their prompt length is between 1,500 and 16,000 tokens
under both the Mistral and Qwen tokenizers; the filtering rule is fixed before
the final model runs.

The candidate set contains FullKV and four compression mechanisms---SnapKV,
PyramidKV, Ada-KV, and L2Norm---at requested retained-token budgets
$B\in\{256,512,1024,2048,4096\}$. This yields 21 configurations and 189
task--configuration cells. All 27,951 request--configuration generations were
completed and scored with the task-specific LongBench metrics.

For a request with $L$ prompt tokens, we report
\begin{equation}
S_{\mathrm{nom}}(r,p)=1-\frac{\min(B_p,L)}{L},
\end{equation}
for compressed configurations, and zero for FullKV. This is a
\emph{nominal retained-token saving}; it is not a measured GPU-memory saving.

\subsection{Request-Level Quality Risk}
Let $Q_{\mathrm{full}}(r)$ be the FullKV score for request $r$ and
$Q_p(r)$ the score produced by configuration $p$. We define non-negative
compression-induced degradation as
\begin{equation}
D(r,p)=\max\left(0,Q_{\mathrm{full}}(r)-Q_p(r)\right).
\end{equation}
For tolerance $\delta$, a violation is
\begin{equation}
Y_{\delta}(r,p)=\mathbb{1}[D(r,p)>\delta].
\end{equation}
Unless noted otherwise, $\delta=0.10$. Across the 26,620 compressed
request--configuration pairs, 16.26\% are violations.

\subsection{Predicting Violations}
The predictor uses information available before compressed candidate execution:
task identity, compression mechanism, requested budget, prompt length,
budget-to-prompt ratio, nominal retention ratio, and inexpensive lexical/shape
features such as character/word counts, line count, punctuation and digit
fractions, and question-mark count. FullKV is excluded from predictor training
because it is the reference configuration.

\begin{figure*}[t]
    \centering
    \includegraphics[width=\linewidth]{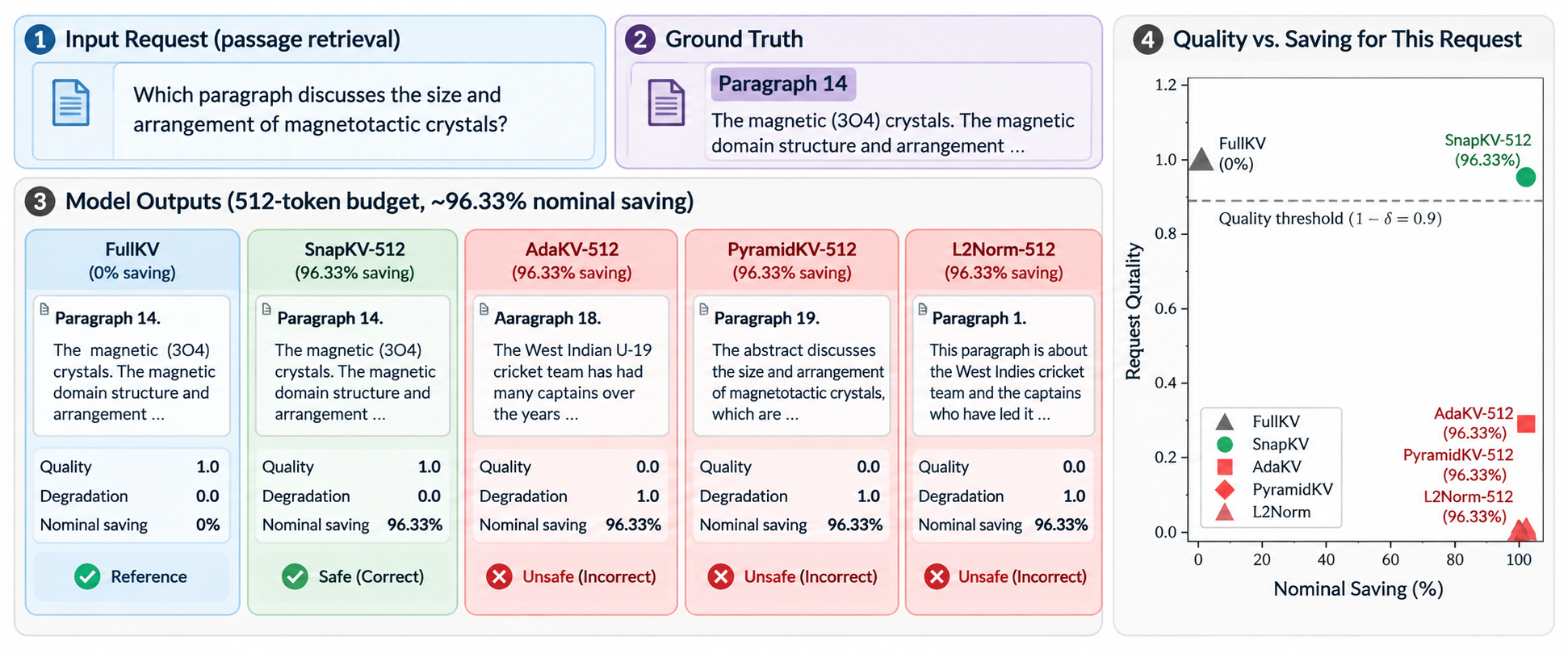}
    \caption{Representative PassageRetrieval-en request at a fixed 512-token budget. FullKV and SnapKV return the correct paragraph, while Ada-KV, PyramidKV, and L2Norm fail despite identical 96.33\% nominal retained-token saving.}
    \label{fig:retrieval}
\end{figure*}

We use a histogram-based gradient-boosted classifier with one-hot categorical
features and standardized numeric features. Evaluation uses five repeats of
five-fold cross-validation grouped by request ID. Consequently, no
configuration of a test request can appear in the corresponding training fold.
We report AUROC, AUPRC, Brier score, negative log-likelihood (NLL), and
15-bin expected calibration error (ECE).

\section{Results \& Discussion}

To illustrate the request-level heterogeneity hidden by aggregate metrics, Fig. ~\ref{fig:retrieval} shows a representative PassageRetrieval-en example evaluated under the same 512-token budget across four KV compression mechanisms. FullKV and SnapKV return the correct paragraph, whereas Ada-KV, PyramidKV, and L2Norm return incorrect passages despite identical nominal retained-token savings of 96.33\%. This example demonstrates that compression budget alone is insufficient to determine quality risk.

\subsection{Method- and Budget-Dependent Safety}
\begin{figure*}[t]
    \centering
    \includegraphics[width=\linewidth]{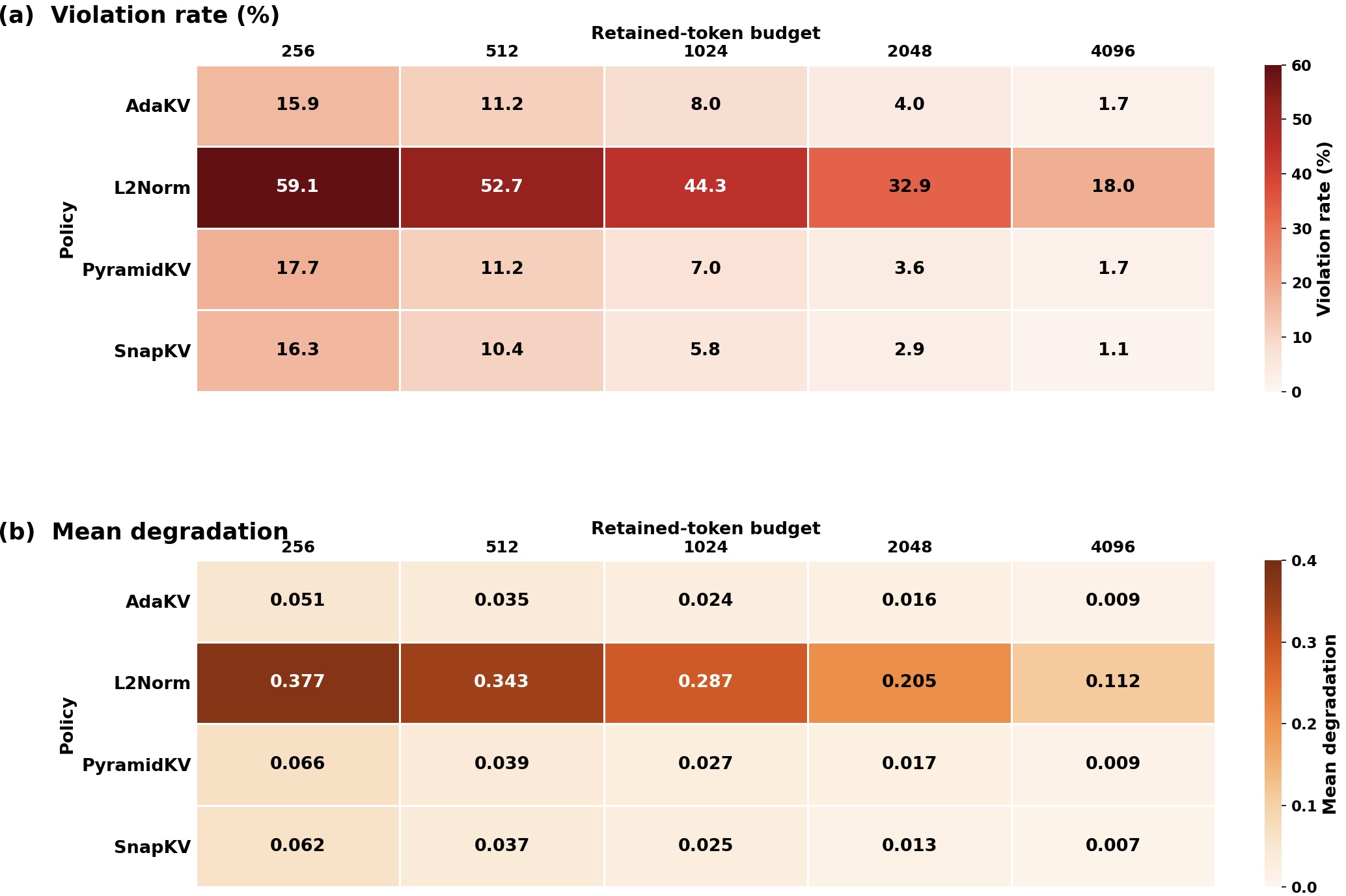}
    \caption{Compression-safety landscape at $\delta=0.10$. Top: request-level
    violation rate (\%). Bottom: mean FullKV-relative degradation. Lower is
    better in both panels.}
    \label{fig:heatmap}
\end{figure*}

The request-level violation landscape in Fig.~\ref{fig:heatmap} shows that
identical retained-token budgets do not imply comparable quality risk. At
4096 retained tokens, violation rates are 1.13\% for SnapKV, 1.65\% for
Ada-KV and PyramidKV, and 18.03\% for L2Norm. At 256 tokens, the rates rise
to 16.30\%, 15.85\%, 17.66\%, and 59.05\%, respectively. Thus, policy
identity materially changes the safety regime even at a fixed token budget.

\begin{figure}[t]
    \centering
    \includegraphics[width=\columnwidth]{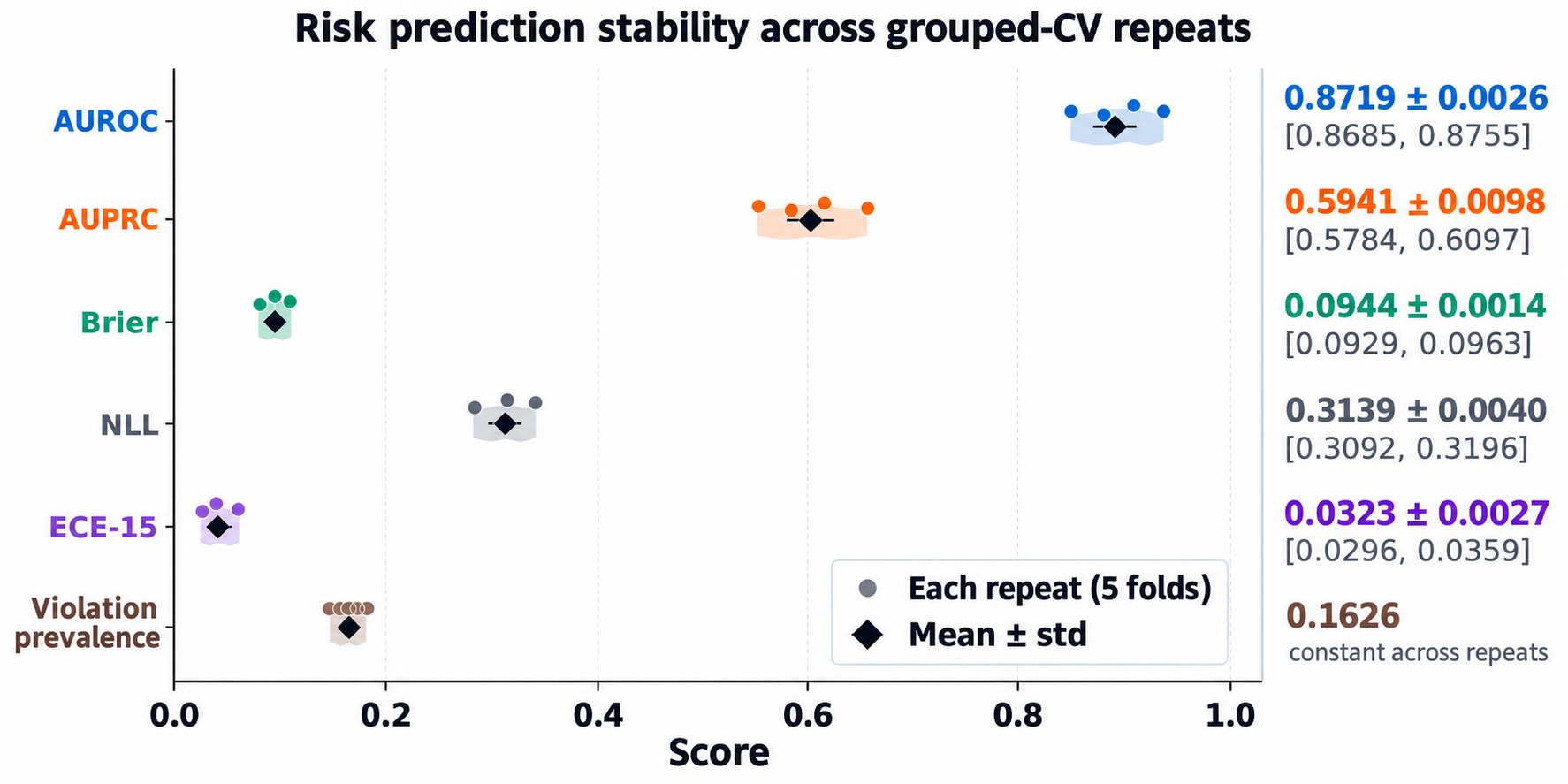}
    \caption{Violation-Risk Prediction at $\delta=0.10$ (mean $\pm$ std over five grouped-CV repeats).}
    \label{fig:cv}
\end{figure}

\begin{table}[t]
\caption{Mean LongBench score by retained-token budget.}
\label{tab:quality}
\centering
\scriptsize
\begin{tabular}{lccccc}
\toprule
Method & 4096 & 2048 & 1024 & 512 & 256\\
\midrule
SnapKV    & .5089 & .5067 & .4978 & .4871 & .4633\\
PyramidKV & .5061 & .5039 & .4943 & .4836 & .4595\\
Ada-KV    & .5090 & .5036 & .4985 & .4869 & .4742\\
L2Norm    & .4113 & .3184 & .2335 & .1745 & .1381\\
\bottomrule
\end{tabular}\\[1mm]
\footnotesize FullKV mean score: 0.5115.
\end{table}

\subsection{Compression Risk Is Predictable}
Fig~\ref{fig:cv} summarizes and visualizes repeated request-grouped validation. AUROC
varies only from 0.8685 to 0.8755 across the five repeats, giving
$0.8719\pm0.0026$. AUPRC is $0.5941\pm0.0098$, substantially above the
0.1626 violation prevalence. The Brier score and ECE indicate useful, though not perfect, probability quality.

\subsection{Oracle Analysis Reveals Large Adaptivity Headroom}
To estimate the maximum benefit available to request-aware execution, we form
a request-wise oracle that chooses the highest-saving candidate satisfying
$D(r,p)\le0.10$, falling back to FullKV only when needed. This oracle attains
94.17\% mean nominal retained-token saving with zero violations and a 0.15\%
FullKV fallback rate. The value is an upper bound rather than a deployable
result: the oracle observes the realized quality of every candidate. Still, it
shows that aggressive compression and low violation rates are not inherently
incompatible; the central challenge is identifying safe candidates before
execution.

As a diagnostic only, an exploratory matched-candidate open-loop quality
predictor selected configurations with 95.57\% mean nominal saving but 15.78\%
violations. We do not present this as a proposed runtime policy; it illustrates
why maximizing predicted quality or compression without explicit risk control
can remain unsafe.

\subsection{Discussion}
Moving on, three implications to consider. \textbf{\textit{First}}, average-quality reporting is insufficient to
describe compression safety: two methods with the same nominal budget can
occupy very different quality regimes. \textbf{\textit{Second}}, the high oracle headroom
indicates substantial per-request heterogeneity rather than a single globally
optimal compression setting. \textbf{\textit{Third}}, the violation predictor shows that this
heterogeneity contains exploitable structure and is not merely random
generation noise. The scope of this study operates currently using one model family (Mistral-7B-Instruct-v0.2) and nine English LongBench tasks. Nominal retained-token saving is an algorithmic quantity; latency, energy, and peak GPU-memory measurements are not claimed here. Simultaneously, the present evaluation here is limited to one model family and does not include a
matched DynamicKV reproduction or additional model-scale checks. These are natural extensions for broader validation. A risk-calibrated resource-aware controller and closed-loop adaptation are also outside the scope of this paper.

\section{Conclusion}
Across 27,951 long-context generations, KV-cache compression exhibits large
request-, mechanism-, and budget-dependent quality variation. At a 0.10
FullKV-relative degradation tolerance, 16.26\% of compressed pairs violate the
quality bound, yet a lightweight predictor identifies risk with
$0.8719\pm0.0026$ AUROC under repeated request-grouped validation. A
request-wise oracle exposes 94.17\% mean nominal retained-token saving with
zero violations, showing that the difficulty is not lack of compression
headroom but safe configuration identification. These findings support
request-level risk as a useful evaluation lens for efficient long-context LLM
inference.

\section*{Acknowledgment \& AI Use Disclosure}
 This work was supported in part by the NYUAD Center for CyberSecurity (CCS), funded by Tamkeen under the NYUAD Research Institute grant G1104.

The authors used Generative AI tools (specifically ChatGPT and Grammarly) for language editing, text refinement, and visual refinement of the methodology figure. The authors reviewed and edited all generated or refined content as needed and take full responsibility for the publication’s content.

\clearpage
\balance
\bibliographystyle{IEEEtran}
\bibliography{refs}
\end{document}